\documentclass[11pt,a4paper]{article}
\usepackage[hyperref]{naaclhlt2019}
\usepackage{times}
\usepackage{latexsym}
\usepackage{float}

\usepackage{url}
\usepackage{graphicx}
\usepackage{amsmath,amssymb}

\usepackage{algorithm}
\usepackage{algcompatible}
\usepackage{algpseudocode}
\usepackage{comment}
\usepackage{booktabs}

\aclfinalcopy 

\definecolor{pink}{RGB}{219, 48, 122}
\definecolor{Orange}{RGB}{255,140,0}
\definecolor{cincinnati-red}{RGB}{190,0,0}

\algrenewcommand\algorithmicrequire{\textbf{Input:}}
\algrenewcommand\algorithmicensure{\textbf{Output:}}

\title{DialectGram: Detecting Dialectal Variation at Multiple Geographic Resolutions}

\author{Hang Jiang\Thanks{ Equal contribution.} \\
  Symbolic Systems \\ 
  \texttt{hjian42@stanford.edu} \\\And
  Haoshen Hong\footnotemark[1] \\
  Computer Science \\
  \texttt{haoshen@stanford.edu} \\\AND
  Yuxing Chen\footnotemark[1] \\
  Symbolic Systems \\
  \texttt{yxchen28@stanford.edu} \\\And
  Vivek Kulkarni \\
  Computer Science \\
  \texttt{viveksck@stanford.edu}}
\date{}
\begin{document}
\maketitle
\begin{abstract}
Several computational models have been developed to detect and analyze dialect variation in recent years. Most of these models assume a predefined set of geographical regions over which they detect and analyze dialectal variation. However, dialect variation occurs at multiple levels of geographic resolution ranging from cities within a state, states within a country, and between countries across continents. In this work, we propose a model that enables detection of dialectal variation at multiple levels of geographic resolution obviating the need for a-priori definition of the resolution level.  Our method \textsc{DialectGram}, learns dialect-sensitive word embeddings while being agnostic of the geographic resolution. Specifically it only requires one-time training and enables analysis of dialectal variation at a chosen resolution post-hoc -- a significant departure from prior models which need to be re-trained whenever the pre-defined set of regions changes. Furthermore, \textsc{DialectGram} explicitly models senses thus enabling one to estimate the proportion of each sense usage in any given region. Finally, we quantitatively evaluate our model against other baselines on a new evaluation dataset \verb|DialectSim| (in English) and show that \textsc{DialectGram} can effectively model linguistic variation.

\end{abstract}

\section{Introduction}

Studying regional variation of language is central to the field of sociolinguistics. Traditional approaches \cite{labov1980locating, milroy1992linguistic, tagliamonte2006analysing,wolfram2015american} focus on rigorous manual analysis of linguistic data collected through time-consuming and expensive surveys and questionnaires. The evolution of the Internet and social media now enables studying linguistic variation at a scale thus overcoming some of the scalability challenges faced by survey based methods. Consequently, computational methods to detect and analyze geographic variation in language have been proposed \cite{eisenstein2010latent,eisenstein2011discovering,eisenstein2014diffusion,bamman2014distributed,kulkarni2015freshman}

However, most prior work suffers from three limitations:  First, previous models \cite{kulkarni2015freshman} such as Frequency Model, Syntactic Model, and \verb|GEODIST| all rely on pre-defined regional classes to model linguistic changes (an exception is \cite{eisenstein2010latent} which focuses on lexical variation). The use of pre-defined regional classes limits the flexibility of these baseline models because dialect changes can be observed at various geographic resolutions. Second, previous models do not explicitly model the sense distribution of each word. In this work, we address these limitations by proposing a model \textsc{DialectGram} that enables analysis at multiple geographic resolutions while explicitly modeling word senses (see Figures \ref{fig:gas} - \ref{fig:pop}). Given a corpus which can be associated with geographical regions, DialectGram first induces the number of senses for each word using a non-parametric Bayesian model \cite{bartunov2016breaking}. This step requires no apriori knowledge of the geographic resolution\footnote{The only requirement is that the corpus be geo-tagged so that analysis can be conducted post-hoc at any desired resolution.}. Having inferred the senses of each word, we show how to detect and analyze dialectal variation at any chosen geographic resolution by clustering usages in any given region based on their sense usage.   

To summarize, our contributions are:
\begin{itemize}
\itemsep0em
    \item \textbf{Multi-resolution Model:} We introduce \textsc{DialectGram}, a method  to study the geographic variation in language across multiple levels of resolution without assuming knowledge of the geographical resolution apriori. 
    \item \textbf{Explicit Sense modeling:} \textsc{DialectGram} predicts how likely each sense of a word is used in a context thus enabling a more precise modeling of linguistic change. 
    \item \textbf{Corpus and Validation Set:} We build a new English Twitter corpus \verb|Geo-Tweets2019| for training dialect-sensitive word embeddings. Furthermore, we construct a new validation set \verb|DialectSim| for evaluating the quality of English region-specific word embeddings between UK and USA.
\end{itemize}

\section{Related Work} \label{sec:related}
\textbf{Linguistic variation}. In the past, sociologists and linguists have been studying linguistic change by designing experiments to manually collect data \cite{labov1980locating, milroy1992linguistic} and conducting variation analysis \cite{tagliamonte2006analysing}. Several works \cite{eisenstein2010latent, gulordava2011distributional, kim2014temporal,jatowt2014framework, kulkarni2015statistically, kulkarni2015freshman, kenter2015ad, gonccalves2016learning, donoso2017dialectometric, lucy2018using, shoemarkroom} have used different computational models to study dialect variations with respective to geography, gender, and time. 

\citet{eisenstein2010latent} is one of the first to tackle the linguistic variation problem with computational models. They design a multi-level generative model that uses latent topic and geographic variables to analyze lexical variation in English. This latent variable model is able to generate an author's geographic location based on the author's text. To quantitatively evaluate the models, they compute the physical distance between the prediction and the true location. Similarly, \citet{gonccalves2016learning} apply $K$-means method to cluster the geographic lexical superdialects assuming a list of pre-defined set of words that are known to demonstrate lexical variation. This was followed by \citet{gonccalves2016learning} who propose two metrics to calculate the linguistic distance between geographic regions. That is, instead of using the physical distance between the predicted and the true location, they compute cosine similarities or Jensen-Shannon Divergence (JSD) to evaluate the model quantitatively. 

Recently, \citet{kulkarni2015freshman} building on the work of \cite{bamman2014distributed} propose a word embeddings based model \verb|GEODIST| model for robustly modeling dialectal variation and focuses on capturing semantic changes between dialects. Nevertheless, a pre-defined set of regions is required for the model to update region-specific embeddings.  For instance, \citet{kulkarni2015freshman} assume that English exhibits dialectal variation between the US and UK, and train the network to learn two sets of word embeddings for the two regions. However, a model trained using this data cannot be used to analyze dialectal variation across states or any other level of resolution without a re-training from scratch. To learn how English changes within each state, \citet{kulkarni2015freshman} would need to tag each US tweet with a state name and train the model again. Moreover, the model does not explicitly capture senses of a word but only learns region specific embeddings.
\textbf{Word Sense Disambiguation}. The problem of detecting dialectal variants of a word can be viewed broadly in terms of word sense induction where the different word senses can roughly correspond to usages in different regions.  For instance, the word \textit{pants} usually refer to \textit{underwear} in the US versus \textit{trousers} in the UK, suggesting two senses for \textit{pants}. Consequently, we discuss the most relevant work on word sense induction as well. \citet{Reisinger2010} is the first paper that modifies the single \textit{prototype} vector space model to obtain multi-sense word embeddings with average cluster vectors as prototypes. Many works \cite{manning2012, McCallum2014, Tian2014APM, Chen2014AUM} are later dedicated to combine Skip-gram, clustering algorithm, and linguistic knowledge to learn word senses and embeddings jointly. \citet{bartunov2016breaking} adopt a non-parametric Bayesian approach and propose the Adaptive Skip-gram (AdaGram) model, which is able to induce word senses without assuming any fixed number of prototypes. As we will see in the following sections, we build on precisely this approach to model regional variation.

\section{Data}
\subsection{Geo-Tweets2019 Corpus}
\begin{table*}[h]
    \centering
    \small
    \begin{tabular}{l||l|l}
        \multicolumn{1}{c||}{\textbf{Word}} & \multicolumn{1}{c|}{\textbf{US Meaning}}              & \multicolumn{1}{c}{\textbf{UK Meaning}} \\ \midrule
        \textit{flat}                               & smooth and even; without marked lumps or indentations & apartment                               \\
        \textit{flyover}                            & flypast, ceremonial aircraft flight                   & elevated road section                   \\
        \textit{pants}                              & trousers                                              & underwear       \\
        \textit{lift}                              & elevator                                              & raise       \\
        \textit{football}                              & soccer                                              & American football       \\
    \end{tabular}
    \caption{Examples of words that have different meanings in American and British English}
    \label{tab:wordlist}
\end{table*}
 We create a new corpus, \verb|Geo-Tweets2019|, which consists of English tweets\footnote{We use the Tweepy toolkit.} during April and May in 2019 from the United States and the United Kingdom. Each tweet includes the user ID, the published time, the geographic location, and tweet text. We have around 2M tweets from the US and 1M from the UK. We preprocessed the tweets with the tweet tokenizer from \citealp{eisenstein2010latent} and regular expressions. Finally, we filtered out URL's, emojis, and other irregular uses of English to shrink the size of vocabulary and to facilitate the training of word vectors. Statistics can be seen in Table \ref{tab:stats}. 


\begin{table}[h]
\small
\centering
\begin{tabular}{c||c|c|c}
\textbf{Number} & \textbf{US}        & \textbf{UK}         & \textbf{Total}     \\ \midrule
tweet  & 2,075,394  & 1,088,232  & 3,163,626  \\
token  & 41,637,107 & 22,012,953 & 63,650,060 \\
term   & 865,784    & 469,570    & 1,167,790 
\end{tabular}
\caption{Statistics of Geo-Tweets2019}
\label{tab:stats}
\end{table}


\subsection{DialectSim Validation Set}

To evaluate the models, we construct a new validation set \verb|DialectSim|, which comprises of words with same or shifted meanings in the US and the UK. To build this validation set, we first crawled a list of words that show different meanings from the Wikipedia page\footnote{\url{https://en.wikipedia.org/wiki/Lists\_of\_words\_having\_different\_meanings\_in\_American\_and\_British\_English}} and pick 341 words that appear more than 20 times in our corpus in the UK and the US. Table \ref{tab:wordlist} presents three examples in the dataset. In order to generate balanced positive and negative samples, we sample another 341 negative examples randomly from our \verb|Geo-Tweets2019| dataset. A minimum frequency of 20 is also used for negative sampling. These negative cases were manually verified by each of the three authors independently. Finally, we split the dataset into training set with 511 samples (75\%) and testing set with 171 samples (25\%). 


\begin{figure*}[h]
\centering
\begin{minipage}{.49\textwidth}
    \centering
    \includegraphics[width=0.985\linewidth]{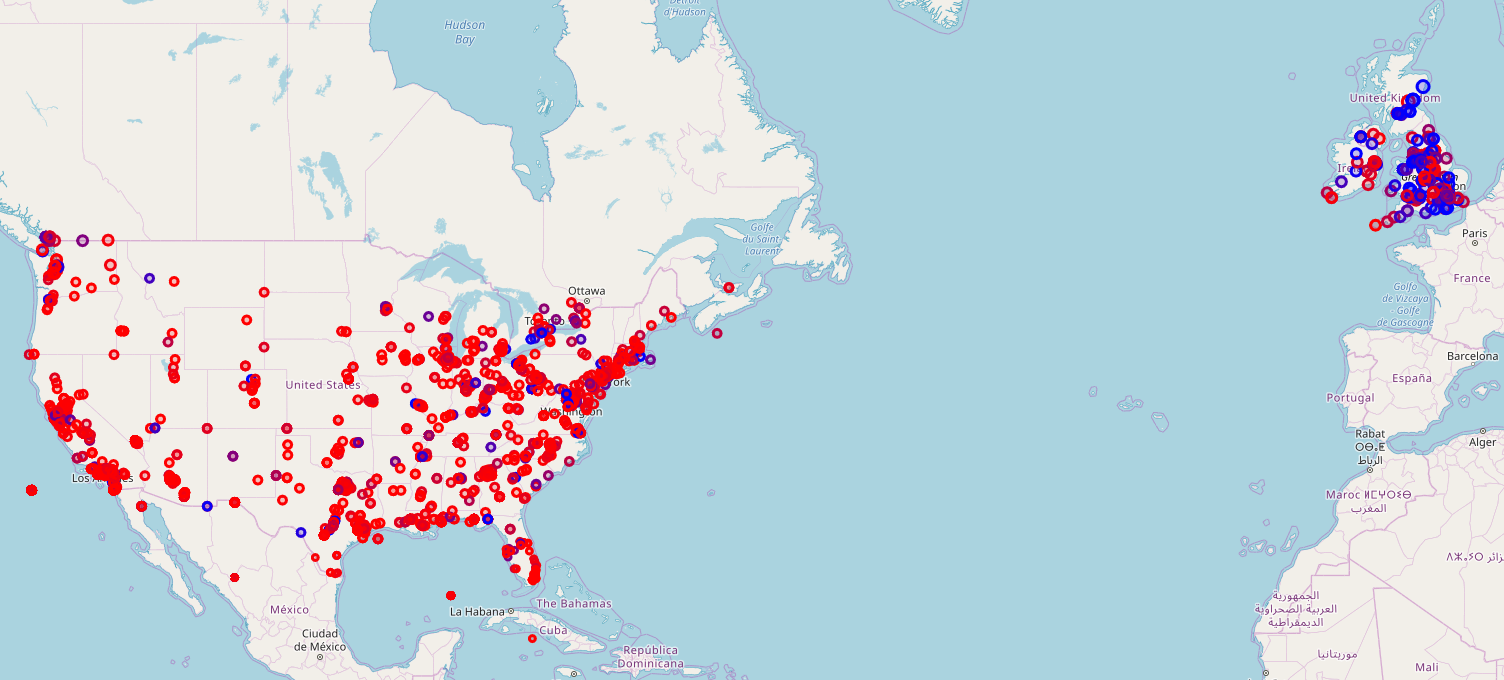}
    \caption{Dialectal variation of \textit{gas} across countries. Tweets that contain \textit{gas} with predicted sense ``gaseous substance'' are illustrated as blue circles; tweets that contain \textit{gas} with predicted sense ``gasoline'' are plotted as red circles.}
    \label{fig:gas}
\end{minipage}%
\hfill
\begin{minipage}{.49\textwidth}
    \centering
    \includegraphics[width=0.95\linewidth]{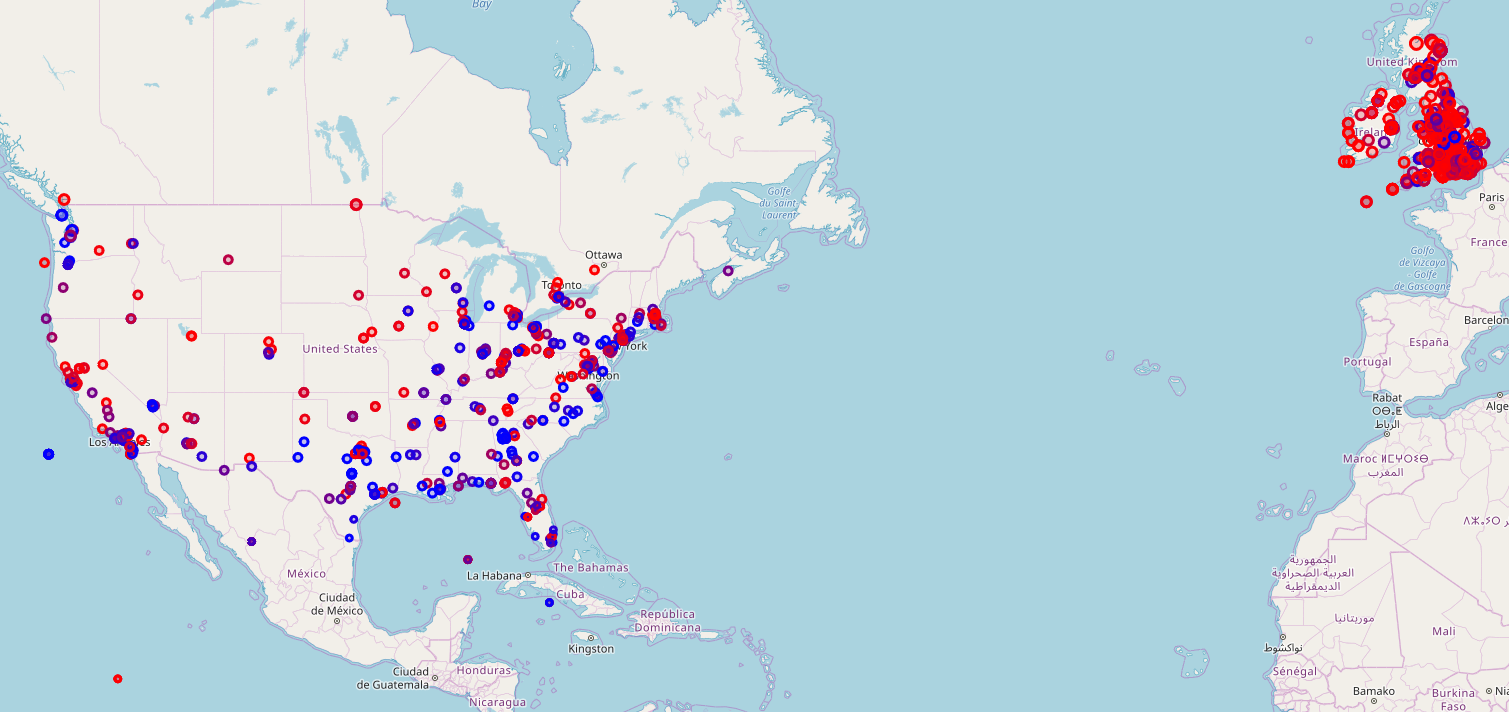}
    \caption{Dialectal variation of \textit{flat} across countries. Tweets that contain \textit{flat} with predicted sense ``apartment'' are illustrated as red circles; tweets that contain \textit{flat} with predicted sense ``smooth and even'' are plotted as blue circles.}
    \label{fig:flat}
\end{minipage}%

\begin{minipage}{.49\textwidth}
    \centering
    \includegraphics[width=0.95\linewidth]{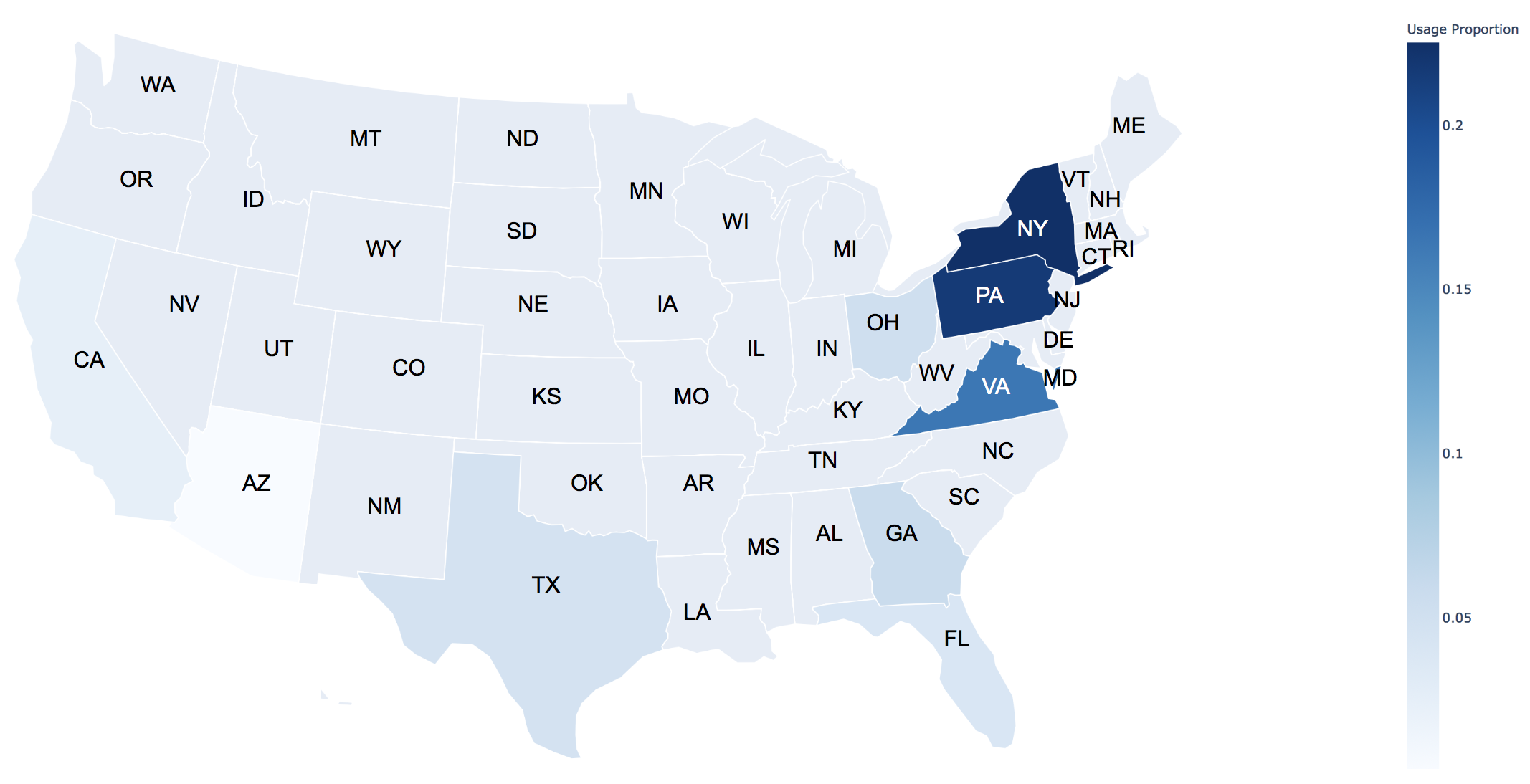}
    \caption{Dialectal variation of \emph{buffalo} across US states. Here we show for each state, the proportion of sense 1 usage (\textit{Buffalo city}) in blue. Grey indicates that the state contains no tweet using the word \textit{buffalo} in our corpus.}
    \label{fig:buffaloState}
\end{minipage}%
\hfill
\begin{minipage}{.49\textwidth}
    \centering
    \includegraphics[width=0.95\linewidth]{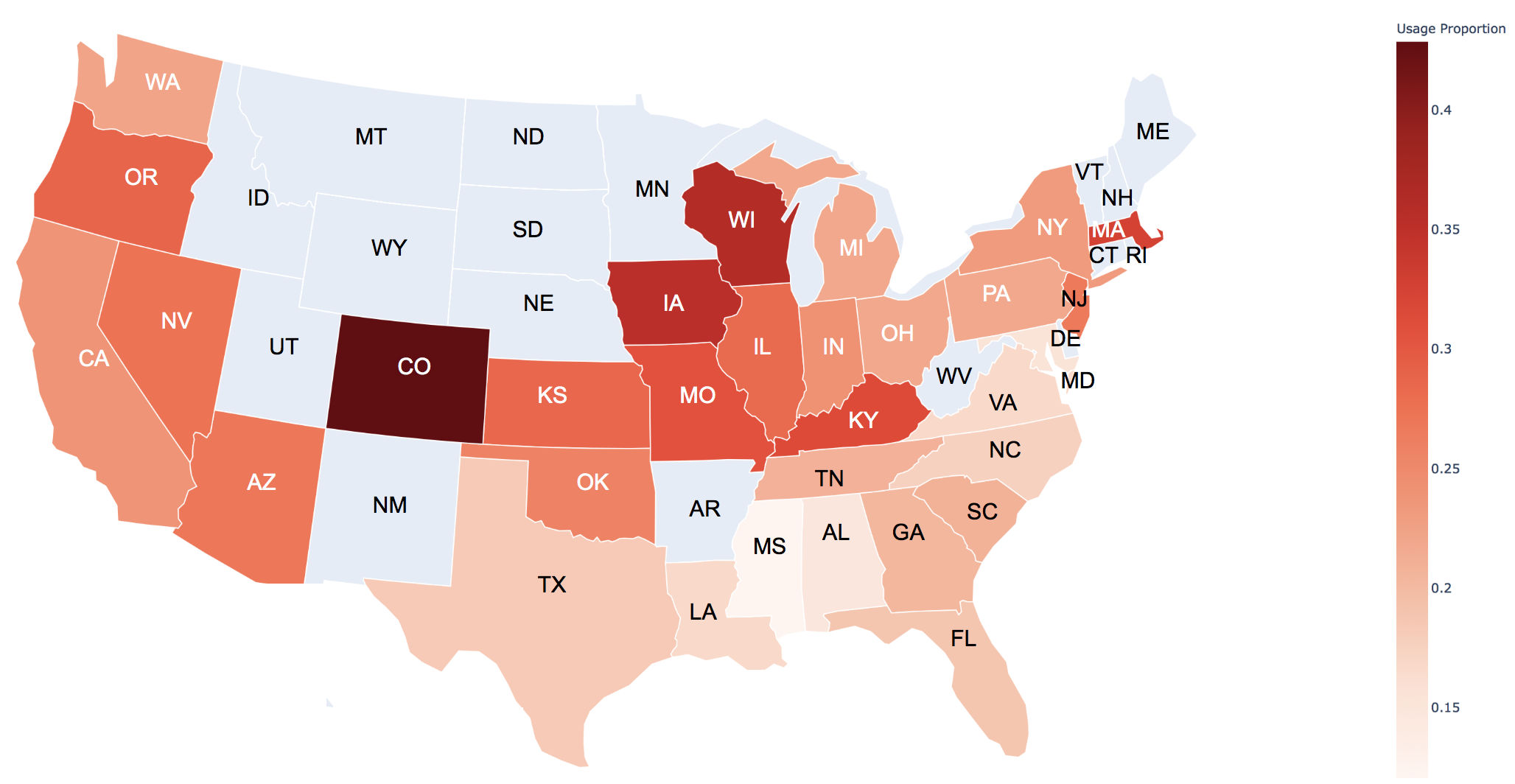}
    \caption{Dialectal variation of \textit{pop} across US states. Here we show for each state, the proportion of sense 2 usage (\textit{soft drink, soda}) in red. Grey indicates that the state contains no tweet using the word \textit{pop} in our corpus.}
    \label{fig:pop}
\end{minipage}
\end{figure*}

\section{Models}
\subsection{Baseline Models}

\textbf{Frequency Model.} One baseline method to detect whether there are significant changes between usage in two regions is to count the occurrence of a word in the US and the UK tweets. We have implemented this Frequency Model as described in \citet{kulkarni2015freshman}.

\textbf{Syntactic Model.} A more nuances approach compared to the frequency based approach is to detect change in syntactical roles across regions. The Syntactic Model \cite{kulkarni2015freshman} takes Part-of-Speech (POS) tag into consideration as well. More specifically, if a word is used equally frequently in both countries, but the their POS usages are different, then we consider the meaning of two words as different between two countries. We use the CMU ARK Twitter Part-of-Speech Tagger\footnote{\url{http://www.cs.cmu.edu/~ark/TweetNLP/}} for POS tagging.  

\textbf{GEODIST (Skip-gram) Model}. The main idea of \verb|GEODIST| model (which can detect semantic changes) \cite{kulkarni2015freshman} is to learn region-specific word embeddings and use boot-strapping to estimate confidence scores on detected changes. Instead of learning a single vector to represent a word, this model aims to jointly learn a global embedding $\delta_{\text{MAIN}}(w)$ as well as (multiple) differential embeddings $\delta_{r_i}(w)$ for each word $w$ in the vocabulary with $R=(r_1, r_2, \dots)$ geographical regions exactly as described in \cite{bamman2014distributed}. In particular, the region-specific embedding is defined as the sum of the global embedding and the differential embedding for that region: $\phi_{r_i}(w) = \delta_{\text{MAIN}(w)} +\delta_{r_i}(w)$. The objective function is to minimize the negative log-likelihood of the context word given the center word conditioned on the region. 
We use stochastic gradient descent method \cite{bottou1991stochastic} to update the model parameters. We implement our own \verb|GEODIST| model in PyTorch. 

\subsection{DialectGram Model}
We construct a new model for detecting dialectal changes which we called \textsc{DialectGram} (Dialectal Adaptive Skip-gram). The model first learns multi-sense word embeddings using Adagram \cite{bartunov2016breaking} through training on the region-agnostic corpus. Once sense specific embeddings are obtained, based on the chosen resolution the model composes region-specific word embeddings by taking a weighted average of sense embeddings. At last, the model calculates the distance between region-specific word embeddings of the same word to determine whether a significant change exists. Our method is described succinctly in Algorithm \ref{alg:DialectGram}. 

Compared to the \verb|GEODIST| model which needs predefined geographic label to update the region-specific embeddings, \textsc{DialectGram} learns multi-sense word embeddings on our dataset without any knowledge of the underlying regions.For instance, DialectGram automatically induces and learns the two senses of the word \textit{flat} which could mean an \textit{apartment} or \textit{level land} corresponding to usages in the UK and US respectively.

\begin{algorithm}[h]
\caption{Use \textsc{DialectGram} to Compose Region-specific Embeddings}\label{alg:DialectGram}
 \begin{algorithmic}[1]
 \Require $w$ word 
 \Ensure $e_r$ weighted region embedding for $w$
 \State Load the trained \textsc{DialectGram} model
 \State Build {$Index_r$} on {$Corpus$} from region $r$
 \For{$s,p\in \Call{GetSensePriors}{w}$} 
    \State {$S_c[s] \gets 0$}, {$S_p[s] \gets p$}
    \Comment{Note: {$S_c$} is sense counts, {$S_p$} is sense priors}
 \EndFor
 \For{all $c \in \Call{GetContexts}{w}$}
    \State{$s \gets \Call{Disambiguate}{w, c}$}
    \State{$S_c[s] \gets S_c[s] + 1$}
 \EndFor
 \State{$e_r \gets \Call{GetWeightedVector}{S_c, S_p}$}
 \end{algorithmic}
\end{algorithm}

\paragraph{Implementation details}
 We train our model on our Geo-Tweets2019 corpus to learn word sense embeddings using the Julia implementation of AdaGram\footnote{\url{https://github.com/sbos/AdaGram.jl}} and then implement the inference algorithm in Python. To obtain a word's region-specific embedding in a place, we first use \textsc{DialectGram} to predict the dominant sense for the word in each tweet from a region and use weighted average of the sense embeddings as the region-specific word embedding $e_r$.  We use the following hyper-parameter settings: $\verb|min_freq| = 20$, $\verb|window_size| = 10$, $\verb|dimension| = 100$, $\verb|maximum_prototype| = 30$, $\alpha = 0.1$, $\verb|epoch| = 1$, $\verb|sense_threshold| = 1e-17$. It is worth noting that a large $\alpha$ (the underlying Dirichlet process) may lead to too many senses for some words and a small $\alpha$, on the contrary, results in too few senses.

To measure the significance of the dialectal change, \citet{kulkarni2015freshman} propose an unsupervised method to detect words with statistically significant meaning changes. However, given that we have access to the humanly curated \verb|DialectSim| dataset, we evaluate the models on the list of annotated words using a simple thresh-holding model (where the thresh-hold parameter is learned from training data). Specifically, We evaluate both Skip-gram models (i.e. \verb|GEODIST| and \textsc{DialectGram}) by calculating the Manhattan distance\footnote{We tried euclidean and cosine distance as well, but use Manhattan distance since it yielded the best results out of the three metrics.} between a word's region-specific embeddings\footnote{Our models, validation set and code are available at: \url{https://github.com/yuxingch/DialectGram}.}.


\begin{table*}[h]
\centering
\begin{tabular}{c||c|c}
\textbf{word} & \textbf{sense 1 neighbors} & \textbf{sense 2 neighbors} \\ \midrule
\textit{gas} & industrial, masks, electric & car, station, bus \\ 
\textit{flat} & kitchen, shower, window & shoes, problems, temperatures \\
\textit{buffalo} & syracuse, hutchinson & chicken, fries, seafood \\
\textit{subway} & starbucks, restaurant, mcdonalds & 1mph, commercial, 5kmh \\
\end{tabular}
\caption{Neighbors of sense embeddings for selected words. This shows \textsc{DialectGram} is able to learn semantic variations of words. }
\label{tab:sense_neighbors}
\end{table*}


\section{Results}
\subsection{Qualitative Analysis} \label{ssec:qualitative}
We investigate the words that \verb|GEODIST| model predicts to have a significant dialectal change between the two regions. For example, the word \textit{mate} is one of the top 20 words in our vocabulary if we sort the vocabulary by the Manhattan distance between the US and the UK embeddings from high to low. However, words like \textit{draft} are predicted to have different regional meanings but not labelled as ``significant'' in \verb|DialectSim|. We further discuss this issue in section \ref{ssec:geodist}.

We select some words with significantly different meanings between the UK and the US. In our \textsc{DialectGram} model, we select the most frequent 2 senses, which usually account for more than 99\% usage variation of a word, and plot a heat map on world map. 


The word maps in Figure [\ref{fig:gas}, \ref{fig:flat}] suggest that the usage of \textit{gas} and \textit{flat} are different in the UK and in the US. \textit{Gas} is used commonly as petrol and related to gas station in the US, but in the UK, \textit{gas} usually refers to air and natural gas. \textit{Flat} could refer to \textit{apartment} but in the US this meaning is not as common as in the UK. The same model can also be used at a different resolution level (across US states). For example, given the word \textit{buffalo}, we show the most dominant senses where \textit{Buffalo City} (in blue) and the \textit{buffalo sauce} sense (in white). Similarly for the word \textit{pop}, we observe that the Midwest area and the Pacific Northwest are more reddish, indicating people are more likely to use the word for \textit{soft drink, soda}, while people in other areas like to use it to describe a certain type of music -- \textit{pop music} \footnote{We normalized the data points by filtering out states where the number of tweets is less than 15 since a small number of data points can suffer from high variance.}.

\subsection{Quantitative Results} \label{ssec:quantitative}

Our training corpus \verb|Geo-Tweets2019| has over three million tweets from US and UK. However, we still observed that micro-level analyses at a resolution lower than the state level required more data samples. Therefore, we only present the country-level and state-level analysis here (note that we do not need to train the model to learn embeddings again when we change resolutions for our analyses).

For each model, we defined a \verb score  function that takes in one word and return a real number denoting its difference in meanings between the UK and the US. We fit a simple threshold model that maximizes the accuracy on training set. Then we test the model performance on testing set. The results are shown in Table \ref{tab:test_performance}. 
\begin{table}[h]
    \centering
    \small
    \begin{tabular}{c||c|c|c|c}
        \textbf{Model} & \textbf{Acc} & \textbf{Prec} & \textbf{Recall} & \textbf{F1}     \\ \midrule
        Frequency                   & 0.5600   & 0.5600    & 0.5887 & 0.5568 \\ 
        Syntactic                   & 0.5263   & 0.5714    & 0.4828 & 0.5233 \\ 
        \verb|GEODIST|              & 0.6432  & \textbf{0.7424}    & 0.5810 & 0.6518 \\ 
        \textsc{DialectGram}          & \textbf{0.6667}   & 0.6837    & \textbf{0.6438} & \textbf{0.6632} \\ 
    \end{tabular}
    \caption{Test performance. Acc, Prec means accuracy and precision. \textsc{DialectGram} has better accuracy, recall, and F1 score than GEODIST.}
    \label{tab:test_performance}
\end{table}
\subsubsection{Frequency Model}
We observed that Frequency Model is more sensitive to word difference between two countries: \textit{football} in the UK is same as \textit{soccer} in the US, causing an imbalanced frequency of term \textit{football} between both countries. However, it can not detect some semantic changes of words if the semantic change preserves frequency for both countries: \textit{flat} has similar frequency in both countries, despite the fact that \textit{flat} could mean \textit{apartment} in the UK, whereas this usage is uncommon in the US. This model does not suffer from an over-fitting problem, because the model is fairly simple and the parameter space is quite small. However the Frequency model is susceptible to a high false positive rate. 

\subsubsection{Syntactic Model}

Syntactic Model performs the worst among all the models. It still gets slightly higher precision than the Frequency Model on test set because it gets some dialectal syntactic changes correct. There are two reasons for its bad performance. First, it is limited by the performance of POS Tagger. Second many word sense changes do not alter POS tags. For example, \textit{pants} refers to \textit{underwear} in the UK while it refers to \textit{jeans} in the US, and both of them are nouns.


\subsubsection{GEODIST Model}\label{ssec:geodist}
As mentioned in Section \ref{ssec:qualitative}, \verb|GEODIST| model is able to detect dialect changes. The accuracy on the test set beats the previous two baseline models (0.6432 versus 0.5600 and 0.5263), as shown in Table \ref{tab:test_performance}. It also outperforms the baseline models in terms of precision and F1 score. In fact,  \verb|GEODIST| model has the highest precision among all models, including the \textsc{DialectGram} model that will be discussed in the next section. We also notice that the recall on the test set is the lowest. The high precision with low recall indicates that for those changes that \verb|GEODIST| model is very conservative and misses some words that actually have significant dialectal changes. For example, the difference between the two region-specific embeddings of the word \textit{pants} is predicted to be not significant, while \textit{pants} does have different meanings in the UK and the US (Table \ref{tab:wordlist}).

\subsubsection{DialectGram Model}
\label{ssec:DialectGram}
DialectGram outperforms the \verb|GEODIST| model in accuracy, recall, and F1 score. However, its precision is lower than that of the \verb|GEODIST| and Frequency Model. However, this is already impressive given the fact that DialectGram does not require pre-determined geographic labels and enables analysis at different geographic resolutions post-hoc (after the model is trained).
One reason for \textsc{DialectGram}'s lower performance in precision compared to \verb|GEODIST| model is that it over-estimates the number of senses (learning senses that overlap).  For example the word \textit{gas} in Table \ref{tab:sense_neighbors}, we sometimes have an additional sense characterized by words such as \textit{air, house, pipe}. This sense seems to be a mix of sense 1, gaseous substance, and sense 2, gasoline. The average number of senses is controlled by  $\alpha$ which we pick based on the model's performance on the training set, but we acknowledge that smarter search strategies for $\alpha$ could be employed.

\section{Conclusion}

In this work, we proposed a novel method to detect linguistic variations on multiple resolution levels. In our new approach, we use \textsc{DialectGram} to train multiple sense embeddings on region-agnostic data, compose region-specific word embeddings, and determines whether there is a significant dialectal variation across regions for a word. In contrast to baseline models, \textsc{DialectGram} does not rely on the region-labels for training multi-sense word embeddings. The use of region-agnostic data allows \textsc{DialectGram} to conduct multi-resolution analysis with one-time training. We also construct \verb|Geo-Tweets2019|, a new corpus from online Twitter users in the UK and US for training word embeddings. To validate our work, we also contribute a new validation set \verb|DialectSim| for explicitly measuring the performance of our models in detecting the linguistic variations between the US and the UK. This validation set allows for more precise comparison between our method (\textsc{DialectGram}) and previous methods including Frequency Model, Syntactic Model, and \verb|GEODIST| model. On \verb|DialectSim|, our method achieves better performance than the previous models in accuracy, recall, and F1 score. Through linguistic analysis, we also found that \textsc{DialectGram} model learns rich linguistic changes between British and American English.
Finally, we conclude by noting the method can be easily extended to temporal or analysis of language at multi-resolution levels.

\section*{Acknowledgments}
We would like to thank Cindy Wang, Christopher Potts, and anonymous reviewers, who gave precious advice and comments to our paper. We would also like to thank Symbolic Systems Program at Stanford University for funding our research through Grants for Education And Research (GEAR). 



\bibliography{naaclhlt2019}
\bibliographystyle{acl_natbib}

\end{document}